\documentclass[runningheads]{llncs}
\usepackage{graphicx}
\usepackage{xcolor}
\usepackage[linesnumbered,ruled,vlined]{algorithm2e}
\SetKwInput{KwInput}{Input}                
\SetKwInput{KwOutput}{Output}              
\usepackage{multirow}
\usepackage{booktabs}
\usepackage{threeparttable}
\usepackage{subfigure}
\usepackage{array}
\usepackage{amsmath}
\usepackage[misc]{ifsym}
\begin{document}
\title{Open-domain Dialogue Generation Grounded with Dynamic Multi-form Knowledge Fusion}
%
%
\author{Feifei Xu\inst{1}
\and Shanlin Zhou\inst{1} 
\and Xinpeng Wang\inst{3}
\thanks{F. Xu, S. Zhou, and X.Wang are the first authors with equal contributions.} 
\and Yunpu Ma \inst{2} $^{(\textrm{\Letter})}$ 
\and Wenkai Zhang\inst{1} 
\and Zhisong Li\inst{4}}
\authorrunning{F. Xu et al.}
%
\institute{
School of Computer Science and Technology, Shanghai University of Electric Power 201306, China \\
\email{xufeifei@shiep.edu.cn,zhoushanlin@mail.shiep.edu.cn}
\and
Chair of Database Systems and Data Mining, University of Munich 80538, Munich\\
\email{cognitive.yunpu@gmail.com}
\and
School of Electronics and Information Engineering, Tongji University 201804, China 
\and
Data Intelligence Center, Alibaba Local Life Service Co., Ltd 200062, China
}
\maketitle              
\begin{abstract}
Open-domain multi-turn conversations normally face the challenges of how to enrich and expand the content of the conversation.
Recently, many approaches based on external knowledge are proposed to generate rich semantic and information conversation. 
Two types of knowledge have been studied for knowledge-aware open-domain dialogue generation: structured triples from knowledge graphs and unstructured texts from documents.
To take both advantages of abundant unstructured latent knowledge in the documents and the information expansion capabilities of the structured knowledge graph, this paper presents a new dialogue generation model, \textbf{D}ynamic \textbf{M}ulti-form \textbf{K}nowledge Fusion based Open-domain \textbf{C}hatt-ing \textbf{M}achine (\textbf{DMKCM}).
In particular, DMKCM applies an indexed text (a virtual Knowledge Base) to locate relevant documents as 1st hop and then expands the content of the dialogue and its 1st hop using a commonsense knowledge graph to get apposite triples as 2nd hop.
To merge these two forms of knowledge into the dialogue effectively, we design a dynamic virtual knowledge selector and a controller that help to enrich and expand knowledge space. 
Moreover, DMKCM adopts a novel dynamic knowledge memory module that effectively uses historical reasoning knowledge to generate better responses. Experimental results indicate the effectiveness of our method in terms of dialogue coherence and informativeness.

\end{abstract}
\section{Introduction}
Open-domain conversation tries to meet human needs in terms of dialogue understanding and emotional resonance while keeping continuous. 
However, traditional merely data-driven multi-turn conversation models often generate simple and repetitive contents \cite{shang2015neural,tang2019target}.
To address this issue, previous studies add additional persona information documents \cite{song2020generating} or guide the conversation topic \cite{xu2020knowledge} to improve dialogue informativeness and diversity.

Notably, more recent studies investigate external knowledge as additional inputs of conversations \cite{zhou2018commonsense,zhang2020grounded,zhao2020multiple,ren2020glks,kim2020sequential}, including knowledge graphs
(denoted as KGs)\cite{zhou2018commonsense,zhang2020grounded,zhao2020multiple}, or unstructured texts \cite{ren2020glks,kim2020sequential}.
Methods based on KGs show that KGs organize information around entities, making it easy to reason. Nevertheless, extracting relations for establishing the knowledge graph usually leads to the loss of information. More, it often generates less informative responses by simply applying and reformulating triples of KGs, e.g., KnowHRL \cite{xu2020knowledge} adds keywords from KGs using reasoning strategies to guide topics but the informativeness of the conversation has not increased significantly.
Informative texts, e.g., comments about movies, can provide rich knowledge for the generation. However, their unstructured representation schemes require the language models to perform knowledge selection or attention from the knowledge texts,e.g., SKT \cite{kim2020sequential} designs a complex screening process to use document knowledge.
In general, these works are impossible to avoid the problem that the KGs are incomplete or the processing of documents is complicated.
A very recent work, MKST \cite{zhao2020multiple} first attempts to apply different forms of knowledge in conversation. 
It extracts the entities mentioned in the sentences and links them to their corresponding entities in KGs as label knowledge. 
It designs a multi-knowledge-aware encoder to encode label, unstructured, and dialogue information together and get a generation by a knowledge-aware decoder. 
However, label knowledge is not achieved through reasoning which may not help the further expansion of dialogue.
More, MKST just relies on dialogue data sets with background knowledge, e.g., Wizard-of-Wikipedia dataset. 

To address above problems, we propose a new multi-turn dialogue generation model, \textbf{D}ynamic \textbf{M}ulti-form \textbf{K}nowledge Fusion based Open-domain \textbf{C}hatting \textbf{M}achine (\textbf{DMKCM}).
Its goal is to fuse abundant knowledge in an indexed corpus (a virtual Knowledge Base or a virtual KB) and information expansion capabilities of a commonsense knowledge graph (commonsense KG) simultaneously to enrich and expand informativeness in multi-turn conversation. 
The differences and functions of these two types of external knowledge can be summarized as follows:
\begin{itemize}
\item 
\textbf{Virtual KB}: This kind of knowledge base is usually an indexed corpus where each document link to its related documents with keywords. Each document in this base can express a complete meaning.
\item 
\textbf{Commonsense KG}: This kind of knowledge graph includes the triples $[head\_entity, relation, tail\_entity]$,  whose entities are also called commonsense facts. These commonsense facts can enhance language representation in the commonsense aspect and even expand topics with reasoning by traversing entities and relations.
\end{itemize}
\begin{figure}
\centering 
\includegraphics[width=\textwidth]{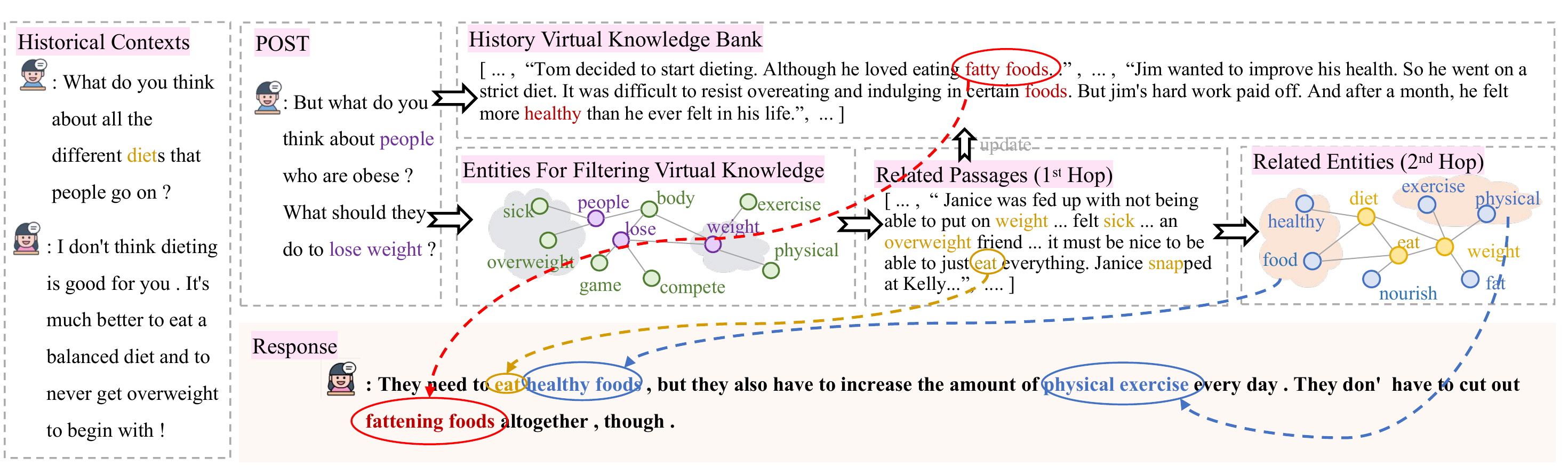} 
\caption{An Example of Knowledge Fusion in a Conversation. Yellow indicate key words from 1st hop, and blue indicate entities of 2nd hop. Red indicates key words from history virtual knowledge. Different colored circles and dotted arrows point out the source of latent knowledge in the response. Black arrows indicate the flow of information.}
\label{fig_example}
\end{figure}

For DMKCM, we design two branches, including a dialogue branch (green blocks in left of Fig.~\ref{fig_overview_detail}) and a knowledge branch (orange blocks in left of  Fig.~\ref{fig_overview_detail}).
The dialogue branch generates responses by interchanging knowledge with the knowledge branch. 
On the knowledge branch,  we separately take different reasoning strategies on virtual KB and commonsense KG to get passages (1st hop) and entities  (2nd hop) that are related to the current dialog.
1st hop services for riching information of response.
2nd hop is to better capture concepts shifted in conversation structure which help generate more meaningful conversations, like “\textcolor[RGB]{232,172,0}{diet}" and "\textcolor[RGB]{232,172,0}{weight}" (concepts in "Historical context" and "POST" in Fig.~\ref{fig_overview_detail}) hop to related concepts, eg, "\textcolor[RGB]{68,114,196}{healthy}" and "\textcolor[RGB]{68,114,196}{exercise}" elt.,  along the commonsense relations, in "Related Entities" of Fig.~\ref{fig_example}. 
This is a typical case in natural conversations.
In addition, before 1st hop,  we also expand concepts by commonsense KG to calculate the filtering scores, like “\textcolor[RGB]{112,48,72}{people}” (from "POST" in Fig.~\ref{fig_overview_detail})  to "\textcolor[RGB]{84,130,53}{overweight}" etl.). Using these filtering scores to select results inferred from Virtual KB helps to remove potential noise in reasoning results.
When the topic is shifted, it is hard to find suitable knowledge from the current 1st hop to generate a response.
Especially,  we find history 1st hop (history virtual knowledge) can solve this issue, e.g., "\textcolor[RGB]{207,62,62}{fatty food}" and "\textcolor[RGB]{207,62,62}{healthy}" in history virtual knowledge bank related to response in Fig.~\ref{fig_example}.
For this,  history virtual knowledge is dynamically stored into the history virtual knowledge bank and provides knowledge support for the current turn. 
This helps the topic transition better in the current dialogue and also enriches the response to some extent.
Our work improves the model explainability on knowledge extraction, helps to generate informative responses, and expands the topic of conversation to a certain extent. 
Explainability is important to dialogue in information-oriented scenarios, where a user needs to know how new knowledge in chatbot’s responses is linked to the knowledge in their utterances, as Fig.~\ref{fig_example} shows.
Our experiments on two conversation datasets, 
including Persona-Chat\cite{zhang2018personalizing} 
and DailyDialog\cite{li2017dailydialog}, demonstrate the effectiveness of DMKCM.

In summary, the following contributions are made in this paper: 
\begin{itemize}
\item
This paper creatively proposes a novel dialogue generation model-DMKCM, to dynamically fuse multi-from knowledge into generation. 
To our best knowledge, this work is the first attempt to fuse virtual KB and commonsense KG into dialogue to get better responses.
\item
To adjust to the open domain dialogue task, we construct a new virtual knowledge base using the dataset of commonsense stories-ROCStory.
\item
We find that history virtual knowledge helps generate better responses and provides a new dynamically delayed updating strategy to store and filter history virtual knowledge.
\item
The experimental results and cases show the superior performance of our model. Various evaluating indicators indicate that DMKCM not only maximizes the advantages of achieved knowledge but also helps to generate more informative and coherent conversations.
\end{itemize}
\section{Related Work}
\subsection{Dialogue Generation with External Knowledge}
Many works have proved that external knowledge can facilitate dialogue generation.
\cite{zhou2018commonsense} presents a novel open-domain dialogue generation method to demonstrate how large-scale commonsense knowledge can facilitate language understanding and generation.
\cite{hayashi2020latent} proposes a latent relation language model, a class of language models that parameterize the joint distribution over the words in a document and relevant entities via knowledge graph relations.
For the use of the external documents, \cite{long2017knowledge} incorporates external documents into the procedure of response generation in custom service dialogues. 
GLKS \cite{ren2020glks} adopts a global guide method to the local, and uses the dialogue contexts to filter out important n-gram information from the document to guide the generation process.
However, the knowledge graphs lose facts, and external texts require complicated processing. These two forms of knowledge still have limitations in exerting external knowledge.
\subsection{Virtual Knowledge base}
Virtual Knowledge Base (virtual KB) is an indexed corpus, which treats a corpus as a knowledge base containing entities and texts. 
It has been widely employed in open-domain Question Answer (QA) \cite{godbole2019multi,moldovan2002performance,Dhingra2020drkit}. Virtual KB accomplishes the QA tasks by answering queries with spans from the corpus, ensuring that facts can be preserved in the relation extraction process.
Whereas, to the best of our knowledge, virtual KB has not yet been mentioned in open-domain dialogue generation.
DrKIT\cite{Dhingra2020drkit} is a state-of-the-art reasoning algorithm with QA on a virtual KB, which traverses textual data like a KB, softly following paths of relations between mentions of entities in the corpus.
Inspired by this, we present a novel model, DMKCM, which includes a reasoning strategy based on DrKIT for getting more information related to our dialogue.
To better fit our task, we convert a commonsense story corpus—the indexed ROCStories\cite{mostafazadeh2016corpus} as our virtual KB, instead of professional Wikipedia. 
\section{Model}
\subsection{Overview}
\begin{figure}
\centering 
\includegraphics[width=\textwidth]{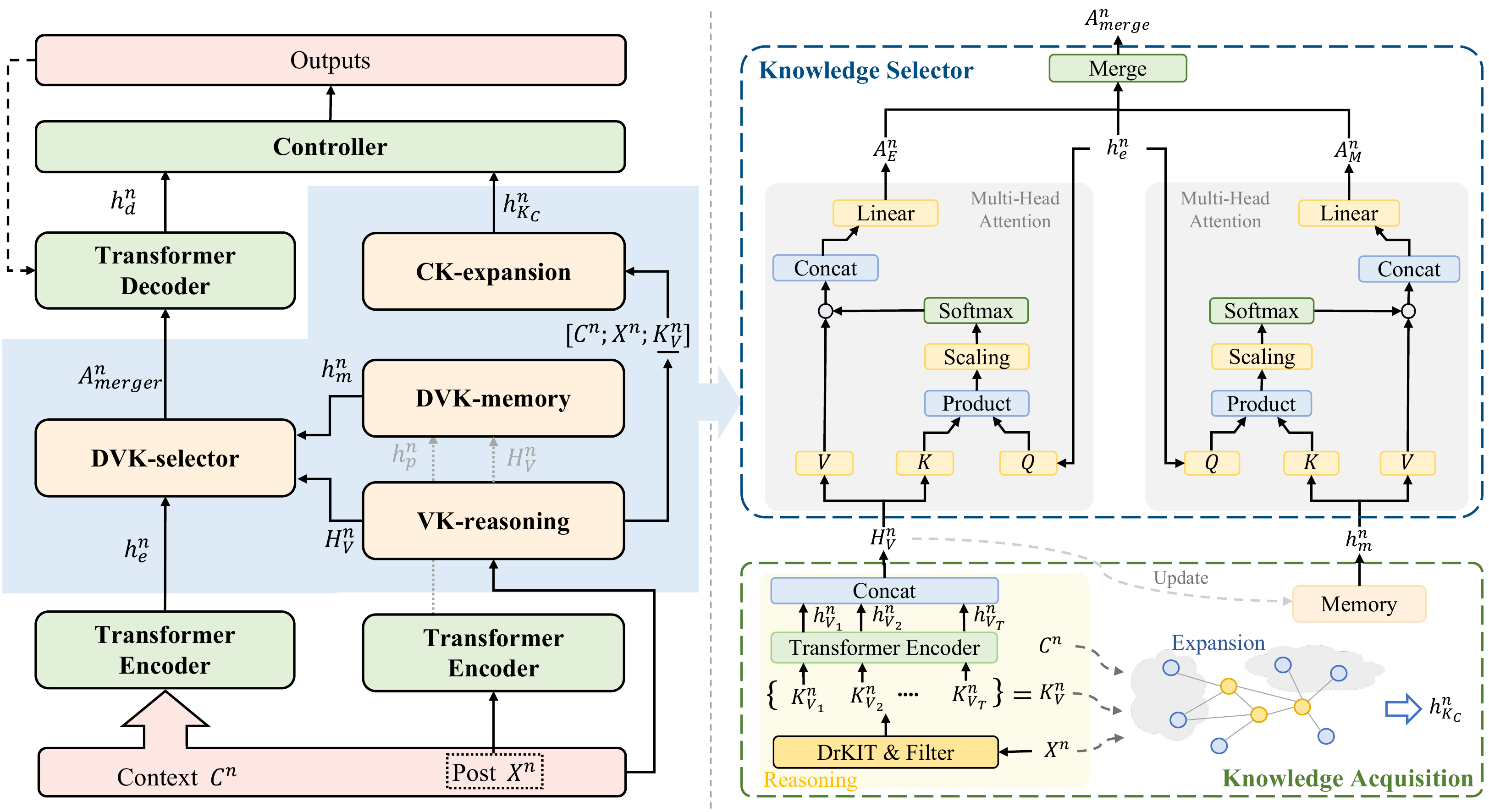} 
\caption{The left is the overview of DMKCM, including the dialogue branch (green blocks) and knowledge branch (orange blocks). Right is details of the knowledge branch, which includes Knowledge Acquisition and Knowledge Selector. Knowledge Acquisition has three modules, including VK-reasoning, DVK-memory, and CK-expansion. Knowledge Selector represents the process of DVK-selector and the merge can be seen in  Eq.\ref{eq_merge}.} 
\label{fig_overview_detail}
\end{figure}

The overview of DMKCM is shown in Fig. \ref{fig_overview_detail}. DMKCM consists of two branches, dialogue branch and knowledge branch. 
\textbf{The dialogue branch }(green blocks in left of Fig.~\ref{fig_overview_detail}) aims to generate conversation based on an encoder-decoder model and interacts information with the knowledge branch  to improve the informativeness expression of response.
\textbf{The knowledge branch}(orange blocks in left of Fig.~\ref{fig_overview_detail}) is to reason, store, merge, and expand knowledge by Virtual Knowledge reasoning module (VK-reasoning), Dynamic Virtual Knowledge memory module (DVK-memory), Dynamic Virtual Knowledge selector module (DVK-selector), and Commonsense Knowledge expansion module (CK-expansion).

Before presenting our detail for the dialogue generation approach, we first introduce our notations and critical concepts.
Formally, suppose we have a conversation dataset
$D=\left\{\left({C}^{i},{X}^{i}, R^{i}\right)\right\}_{i=1}^{N}$, where $C^{i}$ represents conversational context before $i$-th turn. $X^{i}$ is $i$-th user utterance. $R^{i}$ is a response regarding to $X^{i}$. The final goal of this task is to estimate a generation probability distribution
$P(R|[C,X],K)$ from $D$. 
Therefore, one can generate a response for $\left[C,X\right]$ following $P(R|[C,X],K)$, where $\left[C,X\right]$ means the concatenation for context $C$ and current user utterance $X$,  $K$ is the knowledge from knowledge branch, and $R$ is the corresponding response.
\textbf{Assume} that there are already $n-1$ turns in a dialogue.
We use Transformer encoder (T\_enc) to encode $X^{n}$ and $C^{n}$ and get the last hidden state $h_{p}^{n}$ and state $h_{e}^{n}$ from ${X}^{n}$ and $\left[C^{n},X^{n}\right]$ separately.
$h_{p}^{n}$ and $h_{e}^{n}$ represent encoded semantic information of ${X}^{n}$ and $\left[C^{n},X^{n}\right]$.
Next, we elaborate on details for each of them.
\subsection{Knowledge Branch}
Firstly, VK-reasoning reasons and filters candidate documents related to the user utterance $X^{n}$ as 1st hop from a virtual KB. 
We send $X^{n}$, candidate documents, and $C^{n}$ to CK-expansion, aiming to expand commonsense concepts for better response.
DVK-memory is a dynamic transfer module. 
It dynamically stores encoded vectors of 1st hop from previous $n-1$ turns.
We named these vectors as history virtual knowledge.
Then, we filter and send related history virtual knowledge to DVK-selector as a knowledge supplement.
Next, DVK-selector dynamically integrates information from 1st hop, history virtual knowledge and dialog information for the decoder.
Noticeably, after this process, encoded information from the current 1st hop requires to be updated into DVK-memory.  
Before generating a response, CK-expansion needs to expand the words of input information by traversing in a commonsense KG to capture concepts shifted in conversation structure.
These extended concepts are denoted as 2nd hop.
\subsubsection{\textbf{Virtual Knowledge Reasoning(VK-reasoning)}}
Corresponding to our open-domain dialogue task, we select a commonsense story corpus-ROCStories \cite{mostafazadeh2016corpus} as the source of our virtual KB.
Firstly, we index each unique title of ROCStories as an entity in our virtual KB. 
And then, to complete the simulation of the relationship pattern of KGs on the text, we traverse each story to connect related titles (entities).
We use the latest reasoning algorithm DrKIT\cite{Dhingra2020drkit} to train a reasoning model with our conversation dataset and virtual KB. By this trained model, we reason and get the related candidate documents $K_{D}^{n}$ to $X^{n}$.
In particular,  to obtain documents that are more relevant to $X^{n}$, we list related words of each word in $ X ^ {n} $ from a commonsense. 
The number of co-occurrence of each document and this list is regarded as this document's filtering score. 
We get top $T$ candidate documents from $K_{D}^{n}$ by this score.
For convenience, these filtered candidate documents are named as $K_{V}^{n}$ (1st hop).

As shown in right of Fig.~\ref{fig_overview_detail}, we encode $K_{V}^{n}$ for DVK-selector.
Concretely, these candidate documents are successively encoded by transformer encoder. 
Then, we get the last hidden state $h_{V_{t}}^{n}$ from the encoder layer, which represents encoding information of $t$-th candidate document and $t$ is from 1 to $T$. $T$ means the total number of candidate documents. 
Subsequently, we merge the last hidden states generated each time into a matrix called $H_{V}^{n}$. 
The process is as follows:
\begin{equation}
   H_{V}^{n} = \left[h_{V_{1}}^{n},h_{V_{2}}^{n}, \ldots, h_{V_{T}}^{n}\right]^{T},
\end{equation}
\begin{equation}
    h_{V_{t}}^{n} = T\_enc\left(K_{V_{t}}^{n}\right), \left(t=1, \ldots ,T\right).
\end{equation}

\subsubsection{\textbf{Dynamic Virtual Knowledge Memory (DVK-memory)}}
When the topic is shifted, e.g., "\textcolor[RGB]{207,62,62}{\textbf{soda}}" in "A4" of Table~\ref{tab_dd_example}, VK-reasoning may find documents about "\textcolor[RGB]{207,62,62}{\textbf{soda}}" instead of "\textbf{weight}" and "\textbf{healthy}" which stand for the topic from context. 
This leads to little support for the current conversation because our model lacks practical knowledge to generate a response if only using the 1st hop knowledge. 
\begin{table}\scriptsize
\caption{An Example of Conversation Topic Shift in DailyDialog. The topic words in the dialogue are emphasized in bold, and the words in red indicate the words that deviate from the current dialogue topic.}
\resizebox{\linewidth}{!}{ 
\begin{tabular}{p{\textwidth}}
\hline\hline
A1:  What do you think about all the different \textbf{diets} that people go on ? \\[1pt]
B1:  ...a balanced diet and to never get \textbf{overweight}... \\[1pt]
A2:  But...What should they do to \textbf{lose weight} ?  \\[1pt]
B2:  They need to eat \textbf{healthy foods}...cut out \textbf{fattening foods} altogether...\\[1pt]
...\\[1pt]
A4:  How about drinking \textcolor[RGB]{207,62,62}{\textbf{soda}}?  \\[1pt]
B4:   ...\textbf{gain weight} by drinking far too much soda ... \textbf{no nutritional value}..\\[1pt]
\hline\hline
\end{tabular}
}
\label{tab_dd_example}
\end{table}

Thus, we design DVK-memory to dynamically store and filter history virtual knowledge representation, which provides a virtual knowledge supplement for the 1st hop and finally helps the topic transition has better smooth in the current dialogue. This process applies a dynamic delayed updating strategy (in Algorithm \ref{algorithm_DVK-memory}).
\begin{algorithm}
\caption{Algorithm of Dynamically Store and Filter History Virtual Knowledge}
\label{algorithm_DVK-memory}
\DontPrintSemicolon
  \KwInput{The $i$-th (range from 1 to N) turn output of VK-reasoning: $H_{V}^{i}$}
  \KwOutput{Encoded vector of related history virtual knowledge: $h_{m}^{i}$ }
  
    \For{each $i\in [1,N]$}
    {
      \If{the value of $i$ is $1$}
        {
            Do DVK-selector;\; 
            Add $H_{V}^{i}$ into Memory $M^{i+1}$;
        }
        \Else
        {
        	Extract historical knowledge representation of documents $h_{m}^{i}$ related to $h_{p}^{i}$ from $M^{i}$ ;\; 
        	Do DVK-selector;\; 
            Add $H_{V}^{i}$ into Memory $M^{i+1}$;

        }
          
    }
\end{algorithm}

We denote $M=\left\{ M^{n}\right\}_{n=2}^{N}$ as a set of history virtual knowledge representation, where $M^{n}$ is historical knowledge representation for $n$-th turn of dialogue.
\begin{equation}
    M^{n}=\left[H_{V}^{1}, \ldots, H_{V}^{n-1}\right]^{T}.
    \label{formula_Mn}
\end{equation}

Then, We apply attention mechanism from \cite{bahdanau2015neural}  to calculate the extracting historical knowledge representation of documents $h_{m}^{n}$ from $M^{n}$ that is related to the representation of current user utterance $h_{p}^{n}$:
\begin{equation}
  h_{m}^{n}=\sum_{i=2}^{n-1}\alpha _{w,k}^{i}M_{k}^{i},
\end{equation}
\begin{equation}
    \alpha _{w,k}^{i}=\frac{exp\left(S_{w,k}^{i}\right)}{\sum_{i=1}^{n}exp\left(S_{w,k}^{i}\right)},
\end{equation}
\begin{equation}
  S_{w,k}^{i}=V_{a}^{T}tanh\left(W_{h}\left [ h_{p_{w}}^{i};M_{k}^{i} \right ]\right),
\end{equation}
where $M_{k}^{i}$ represents the k-th position hidden state of history virtual knowledge $M^{i}$ and $h_{p_{w}}^{i}$ is the w-th token vector in $h_{p}^{i}$. $V_{a}^{T}$ and $W_{h}$ are trainable parameters. $S_{w,k}^{i}$ is the unnormalized attention weight by an attention neural network and $\alpha _{w,k}^{i}$ is the normalized attention weight from $S_{w,k}^{i}$.
\subsubsection{\textbf{Dynamic Virtual Knowledge Selector(DVK-selector)}}
We apply multi-head attention mechanism (MultiHead) \cite{vaswani2017attention} to extract features of current
virtual knowledge $H_{V}^{n}$ and historical knowledge $h_{m}^{n}$ according to dialogue semantic information $h_{e}^{n}$. 
A gate is proposed for information fusion and its result is $A_{merge}^{n}$.
Specifically, 
\begin{equation}
\begin{split}
 A_{merge}^{n}= \left \{
 \begin{array}{ll}
    \mu A_{V}^{n}+h_{e}^{n}, & n=1\\
     \mu A_{V}^{n}+(1-\mu)A_{M}^{n}+h_{e}^{n}, & n > 1 \\
 \end{array}
\right.
\end{split},
\label{eq_merge}
\end{equation}
\begin{equation}
    A_{M}^{n}=MultiHead\left(h_{e}^{n}, h_{m}^{n}, h_{m}^{n}\right),
\end{equation}
\begin{equation}
    A_{V}^{n}=MultiHead\left(h_{e}^{n}, H_{V}^{n}, H_{V}^{n}\right),
\end{equation}
\begin{equation}
   \mu = sigmoid\left(W_{g}h_{e}^{n}\right).
\end{equation}
Here, we use the sigmoid function to get a gating parameter $\mu$ for fusion, and the $W_{g}$ are trainable parameters.
$ A_{V}^{n}$ is the current virtual knowledge features related to $h_{e}^{n}$ and $A_{M}^{n}$ is the historical knowledge features related to $h_{e}^{n}$.
Particularly, since DVK-memory takes a delayed updating strategy, it needs to remove the $h_{m}^{n}$ when $n$ is 1.
\subsubsection{\textbf{Commonsense Knowledge Expansion(CK-expansion)}}
To expand concepts and further enhance informativeness, we expand entities of $K_{V}^{n}$, $C^{n}$ and $X^{n}$ by searching neighbor nodes on a commonsense KG.  
We use $K_C^{n}=(k_h^{n},k_r^{n},k_t^{n})$ to represent the knowledge triples, which connects the original entities and expanded entities. 
$k_h^{n}$ is a set of words (entities) from $K_{V}^{n}$, $C^{n}$ and $X^{n}$.
$k_t^{n}$ means expanded entities by the KG.
$k_r^{n}$ is the relation of $k_h^{n}$ and $k_t^{n}$ on the KG.
Inspired by GCN that can encode graph structure well, we use Multi-layer CompGCN (M\_CompGCN) \cite{vashishth2020compositionbased} to encode the knowledge triples by combining the node embedding and the relation embedding.
\begin{equation}
    h_{K_h}^n, h_{K_r}^n, h_{K_t}^n = M\_CompGCN(K_C^{n}).
\end{equation}
We use the dialogue context encoding $h_e^n$ to compute the degree of attention $\beta^i$ with the encoded head $h_{K_h}^n$ and the relation $h_{K_r}^n$ , and then multiply with the encoded tail $h_{K_t}^n$. Finally, we get the representation of  knowledge triples $h_{k_C}^n$.
\begin{equation}
    h_{k_C}^n = \sum_{i=1}^k{\beta^i h_{k_t}^i},
\end{equation}
\begin{equation}
    \beta^i = Softmax(h_{e}^i[h_{k_h}^i+h_{k_r}^i]),
\end{equation}
where k is the number of the triples.
\subsection{Generation}
We use Transformer Decoder (T\_dec) to generate words,
\begin{equation}
    h_{d}^{n} = T\_dec(y_{t-1}^{n}, A_{merge}^{n}).
\end{equation}
Then, a Controller is designed, in which the decoded hidden state $h_{d}^{n}$ will be mapped into vocab size and outputs the probability of words $P_{v}$ by Softmax function,
\begin{equation}
    P_{v} = Softmax(W_{v}h_{d}^{n}).
\end{equation}
In addition, we can also generate knowledgeable words by using knowledge expansion representation encoded in CK-expansion,
\begin{equation}
    P_{K_{C}} = Softmax(\sum_{i=1}^l{\gamma_i^n h_{k_C}^n}),
\end{equation}
\begin{equation}
    \gamma_i^n = Softmax(h_{d}^{n}W_{k}h_{K_C}^n).
\end{equation}
We get an attention weight $\gamma_i^n$ by using the decoded hidden state $h_{d}^{n}$ to focus on the $h_{k_C}^n$, which can make the model focus on the relative knowledge triples; then, we choose the knowledge entities according to entities probability $P_{K_{C}}$ of relative weighted knowledge after Softmax function.

The final generated words will consider both the distribution of standard vocabulary and the distribution of knowledge entities.
We use a soft gate probability $g_t$ to choose the generated words from standard vocabulary or knowledge entities.
\begin{equation}
    y_{t} = g_{t} \cdot P_{v} + (1-g_{t}) \cdot P_{K_{C}}
\end{equation}
\begin{equation}
    g_{t} = \sigma(h_{d}^{n})
\end{equation}
\subsection{Training}
To train the proposed model, we minimize the negative log-likelihood
\begin{equation}
    L_{NLL} = - \frac{1}{N} \sum_{i=1}^{N}\sum_{t=1}^{T}log P(y_{t}^{(n)}|y_{<t}^{(n)},X^{(n)}, K^{(n)}),
\end{equation}
where $N$ is the total number of the dataset, and $T$ is the timestep of the $n$-th turn response sentence. $X^{(n)}$ represents the $n$-th turn user utterance in the dataset, and $K^{(n)}$ represents the $n$-th turn knowledge.
\section{Experiments}
\subsection{Dataset}
\begin{figure}
    \centering
    \includegraphics[width=\textwidth]{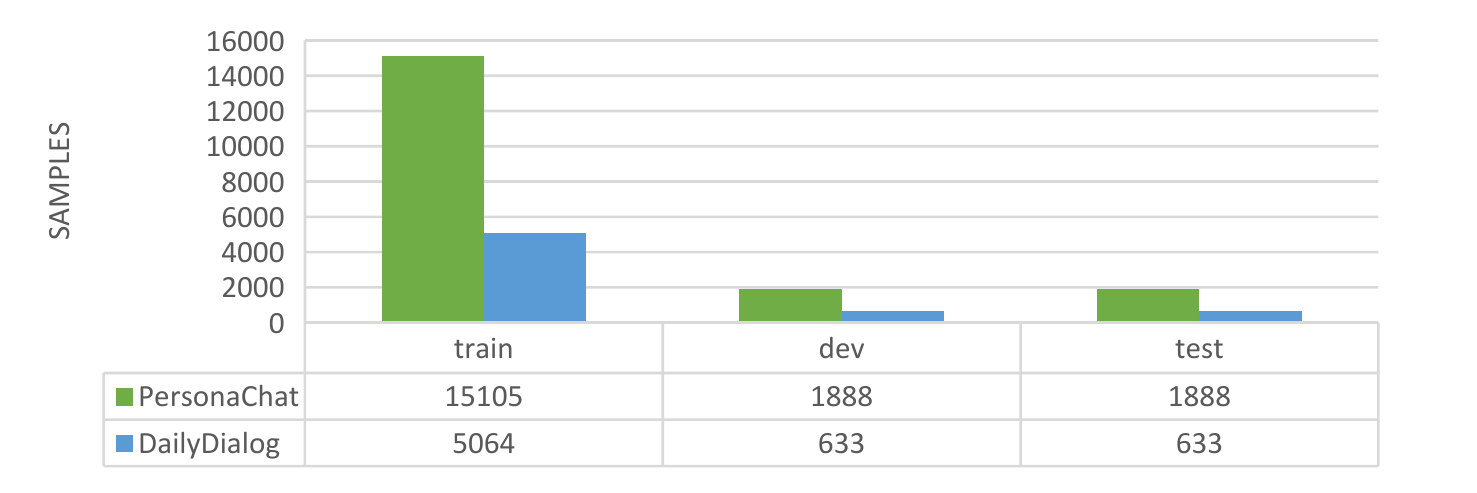} 
    \caption{Statistics Of Two Datasets.}
    \label{fig_datasets}
\end{figure}

\textbf{Conversation Corpus:} 
We choose DailyDialog \cite{li2017dailydialog} 
and PersonaChat \cite{zhang2018personalizing} 
as our datasets.
In our work, four turns of dialogue are a unit of training sample and pre-processed statistics of the above datasets are shown in the Fig.~\ref{fig_datasets}.
\textbf{Commonsense Knowledge Corpus: } ConceptNet\cite{speer2017conceptnet} is a semantic network designed to help computers understand the meanings of words that people use.
Its English vocabulary contains approximately 1,500,000 nodes. 
\textbf{Source of Virtual KB: } 
The ROCStories \cite{mostafazadeh2016corpus} is a commonsense story corpus 
which contains 98,161 five-sentence stories.
\subsection{Comparison Method}
We compare our model with representative baselines to investigate its effectiveness. The baselines are as follows:
\textbf{(1) Attn-S2S}\cite{sutskever2014sequence}: 
A classic method applies simple attention \cite{bahdanau2015neural} to the input context based on the sequence-to-sequence model;
\textbf{(2) Transformer}\cite{vaswani2017attention}: 
Transformer is a popular network architecture, based solely on attention mechanisms;
\textbf{(3) Dir-VHRED}\cite{zeng2019dirichlet}:
A recent work based on the latent variable hierarchical recurrent encoder-decoder model and characterizes the latent variables using Dirichlet distribution instead of traditional Gaussian distribution;
\textbf{(4) GLKS}\cite{ren2020glks}:
The newest dialogue generation model based on unstructured knowledge builds a global-to-local knowledge selection method to improve the quality of selected unstructured knowledge in background-based conversations;
\textbf{(5) CCM}\cite{zhou2018commonsense}:
A commonsense knowledge-aware conversation model, which leverages commonsense knowledge from ConceptNet\cite{speer2017conceptnet} through two graph attention mechanisms to facilitate informative response generation;
\textbf{(6) MKST}\cite{zhao2020multiple}: 
The latest universal transformer-based architecture fuses label, background knowledge in open-domain conversation. 
Since MKST relies on datasets with background knowledge, we compare it with our model only on PersonaChat which has background information.
\subsection{Implementation Details}
We conduct the experiments with a Transformer structure (our baseline) with 8 heads, 6 layers, 512-dimensional hidden states, and a 2-layer GCN model.
In the VK-reasoning, we set the number of reasoned candidate documents as 10 and filtered candidate documents as 5.
During the CK-expansion, we search the neighbors of nodes and preserve the top 100 neighbor nodes.
When processing datasets, the history context we choose are the previous turns of the current conversation.
To train the model, we use the Adam optimizer\cite{kingma2014adam} and use Adam-warmup to adjust the learning rate.
\subsection{Evaluation Metrics}
To analyze and evaluate our model more comprehensively, we use both automatic and human evaluations.
\textbf{Automatic Evaluation}:
Based on previous work, we apply several widely used automatic metrics.
Specifically, we adopt PPL,BLEU-{1,2,3,4})\cite{papineni2002bleu},and  Distinct-{1,2} (Dist-{1,2})\cite{li2016diversity} to intuitively reveals quality, coherence and diversity of generated responses.
PPL is the perplexity score that measures the quality of the language model.
BLEU calculates word-based precision between a generated response and a gold response. Distinct evaluates the informativeness and diversity of the predicted responses. 
\textbf{Human Evaluation}:
As known, automatic evaluation indicators have limitations in evaluating human conversations \cite{Liang2021HERALDAA}.
In our work, we randomly sample 200 test samples to conduct human evaluations. 
For response, we define three metrics: 
(1) Fluency (\textbf{Flu.}), i.e., degree of fluency and human readability; 
(2) Informativeness (\textbf{Inf.}), i.e., degree of knowledge for responses;  
(3) Appropriateness (\textbf{App.}), i.e., degree of relevance to the given context;  
Each response has 3 annotators to give a 3-graded whose rating range from 0 to 2. We take the average scores as the final results for each metric.
\section{Results and Discussion}
\subsection{Performance Evaluation}
\subsubsection{Automatic Evaluation}
Table~\ref{tab_comp} lists the automatic evaluation results for each model. Our model outperforms almost the baselines on two corpora. 
In the quality of the model, our PPL is the lowest, indicating that our generated responses Model is more grammatical. 
In the aspect of coherence, DMKCM has higher BLEU values, demonstrating our model tends to generate responses that are more similar to the gold responses than baselines in most cases. 
It can be inferred that our model can effectively obtain useful information from the historical context and historical knowledge in memory to help generate a response. 
On diversity, the Dist-1,2 metrics demonstrate that the models leveraging external knowledge achieve better performance than the knowledge-based model, e.g., GLKS, CCM, MKST, in generating meaningful and diverse responses.
According to Table~\ref{tab_comp}, in terms of indicators, DMKCM is better than MKST,  which is the latest multi-knowledge based dialogue generation model.
This signifies the effectiveness of our model on using structured knowledge or unstructured knowledge or the method of fusion.
\begin{table}
\centering
\caption{Automatic Evaluation Results of the Proposed Model and the Baseline Models. Numbers in bold indicate the best-performing model on the corresponding metrics.}
\resizebox{\linewidth}{!}{ 
\renewcommand{\arraystretch}{0.8}
\begin{tabular}{cllllllll}
\toprule
\textbf{Dataset} & \textbf{Model} & \textbf{PPL} & \textbf{BLEU-1} & \textbf{BLEU-2} & \textbf{BLEU-3} & \textbf{BLEU-4} & \textbf{Dist-1} & \textbf{Dist-2} \\
\midrule
\multirow{7}{*}{\textbf{PersonaChat}} 
 & Attn-S2S & 7.0079 & 0.4372 & 0.2525 & 0.1458 & 0.0842 & 0.0185 & 0.1208\\
 & TRANSFORMER & 6.5172 & 0.5023 & 0.2900 & 0.1675 & 0.0967 & 0.0425 & 0.2239 \\
 & Dir-VHRED & 8.7117 & 0.4428 & 0.2557 & 0.1476 & 0.0852 & 0.0164 & 0.0795 \\
 & GLKS & 7.5668 & 0.4684 & 0.2705 & 0.1562 & 0.0903 & 0.0420 & 0.1556 \\
 & CCM & 7.3534 & 0.4703 & 0.2715 & 0.1568 & 0.0906 & 0.0596 & 0.2373 \\
 & MKST & 7.1307 & 0.4460 & 0.2577 & 0.1491 & 0.0864 & 0.0808 & 0.2907   \\
\cmidrule{2-9}
& \textbf{DMKCM} (our model) & {\textbf{5.4549}} & {\textbf{0.5452}} & {\textbf{0.3149}} & {\textbf{0.1852}} & {\textbf{0.1087}} & {\textbf{0.0903}} & {\textbf{0.3328}} \\
\midrule\midrule
\multirow{6}{*}{\textbf{DailyDialog}} 
 & Attn-S2S & 7.7451 & 0.4166 & 0.2406 & 0.1390 & 0.0803 & 0.0281  & 0.2010  \\
& TRANSFORMER & 10.1355 & 0.4205 & 0.2428 & 0.1403 & 0.0811 & 0.0633  & 0.3096 \\
& Dir-VHRED & 9.3176 & 0.4276 & 0.2469 & 0.1426 & 0.0824 & 0.0399 & 0.2052 \\
& GLKS & 7.7635 & 0.4449 & 0.2570 & 0.1485 & 0.0859 & 0.0738 & 0.3496 \\
& CCM & 7.5592 & 0.4875  & 0.2773  & 0.1560  & 0.0860 & 0.0598 & 0.2350  \\
 \cmidrule{2-9}
& \textbf{DMKCM} (our model) & {\textbf{5.7692}} & {\textbf{0.4928}} & {\textbf{0.2846}} & {\textbf{0.1645}} & {\textbf{0.0951}} & {\textbf{0.0801}} & {\textbf{0.3589}}\\
\bottomrule
\end{tabular}}
\label{tab_comp}
\end{table}
\subsubsection{Human Evaluation}
\begin{figure}
\centering
\subfigure[Results of PersonaChat]{
\begin{minipage}[t]{0.45\linewidth}
\centering
\includegraphics[scale=0.36]{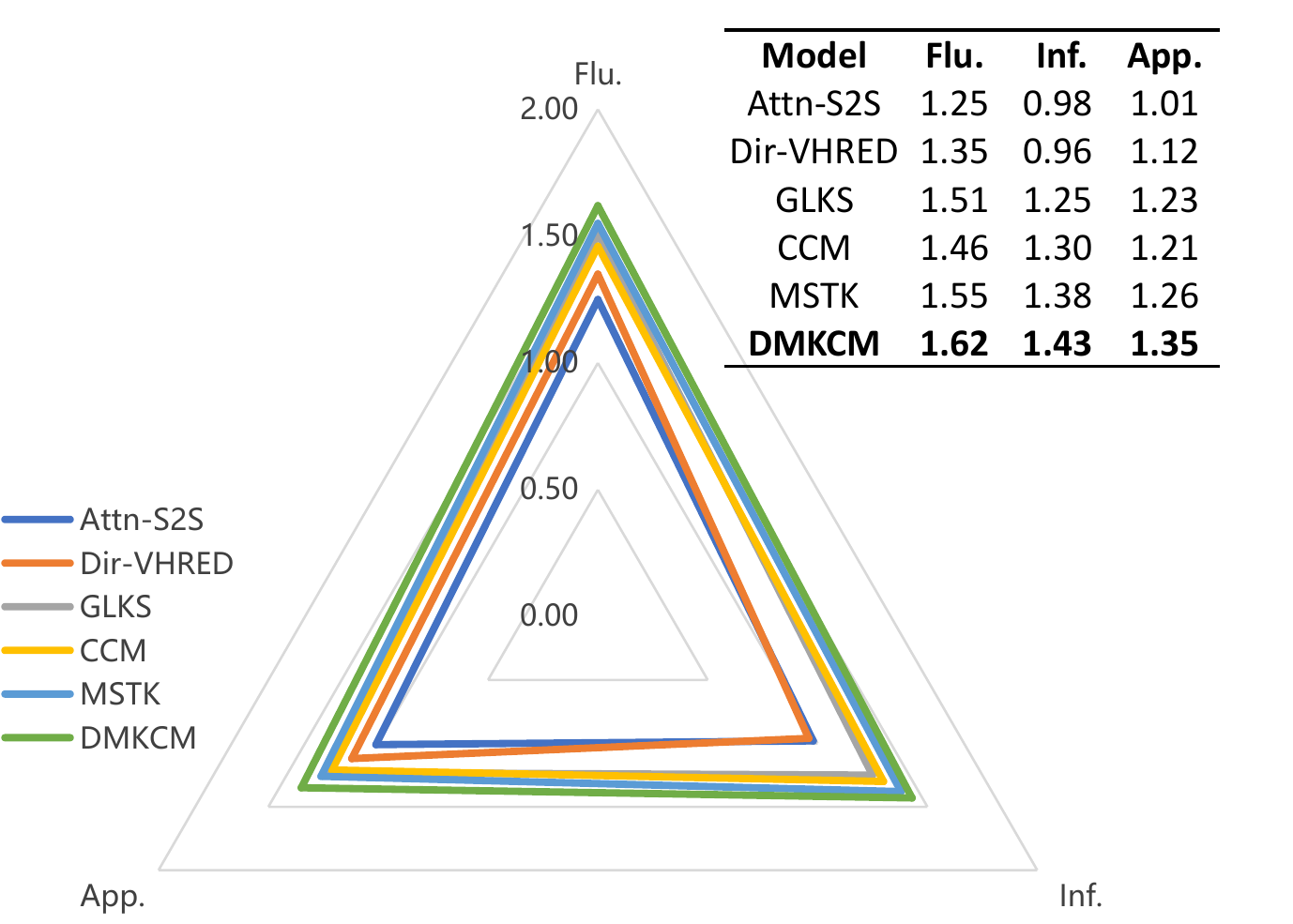} 
\end{minipage}
}
\subfigure[Results of DailyDialog]{
\begin{minipage}[t]{0.45\linewidth}
\centering
\includegraphics[scale=0.35]{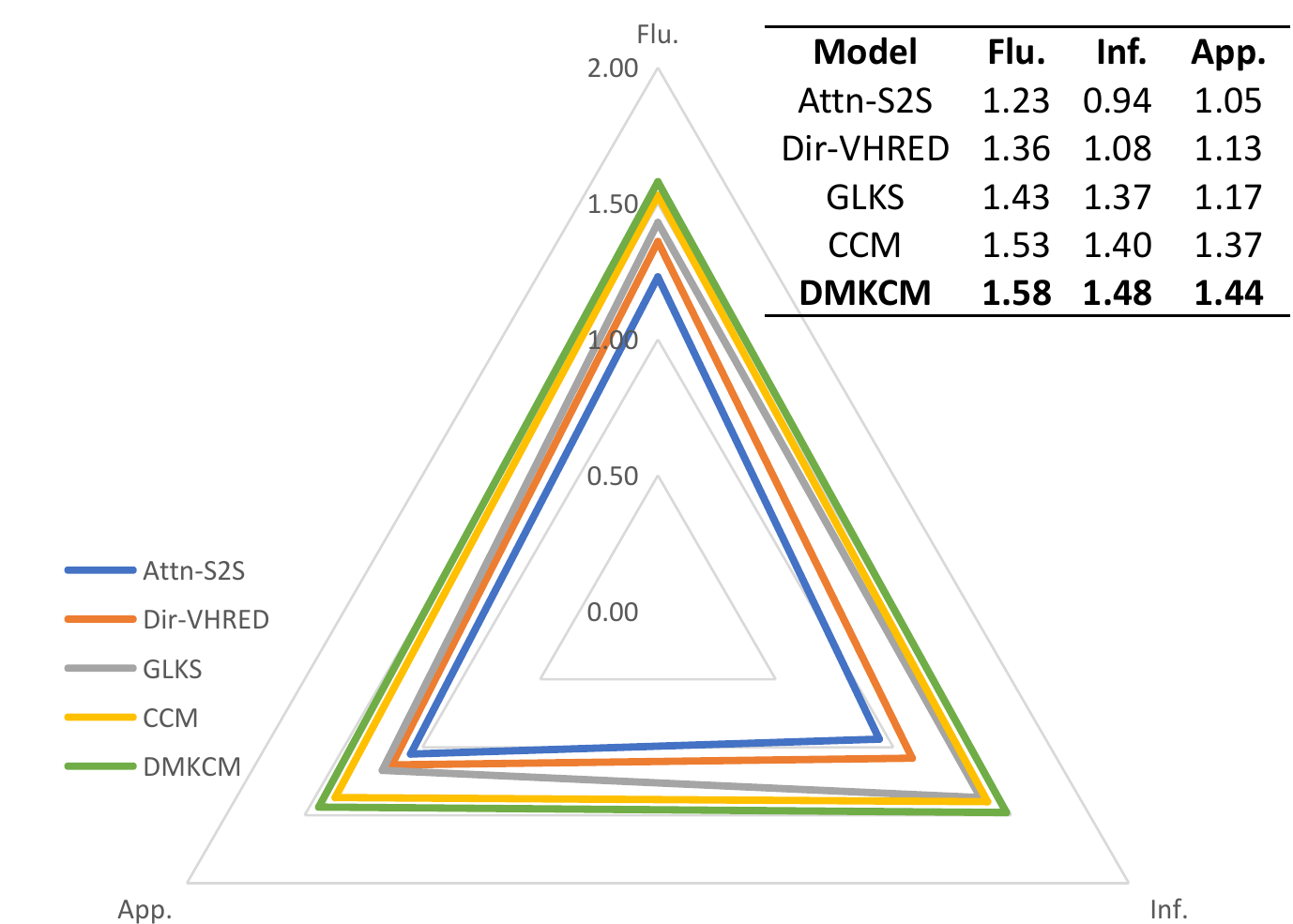}
\end{minipage}
}
\centering
\caption{Comparison of Human Evaluation Results.The rating range from 0 to 2 and the bigger means the better.}
\label{human}
\end{figure}

Fig.~\ref{human} clearly shows the human evaluation metrics results of DMKCM compared with the baselines through the radar chart. 
The three vertices of the radar chart respectively represent fluency, informativeness, and appropriateness. From the radar chart, 
DMKCM has the best performance on two datasets. 
Particularly, the informativeness has the most obvious advantage over other baselines, which indicates the effectiveness of our fusion of the multi-form knowledge and can generate coherent and informative responses.
\subsection{Ablation Study}
As shown in Table~\ref{tab_abl}, we analyze the effectiveness of each module proposed in DMKCM through the following situations:
\textbf{(1)- 2Hop}: DMKCM drops CK-expansion;
\textbf{(2)- Mem and 2Hop}: DMKCM drops DVK-memory and CK-expansion;
\textbf{(3)- 1Hop and Mem}: DMKCM drops VK-reasoning and DVK-memory;
\textbf{(4)- 1Hop, Mem, and 2Hop}: This is baseline (Transformer).
From the results,  we can observe that the performance of situation (1) drops sharply from our model. 
This is within our expectations since CK-expansion helps to capture extra information from the post, which can improve the diversity of generated responses. 
This result can also show that fusing structured knowledge can effectively help dialogue generation.
Metric results of the situation (1) are better than situation (2), which verifies that retaining history virtual knowledge using DVK-memory can effectively help dialogue generation.
Situation (3) is related to unstructured knowledge and its poor results prove the effectiveness of VK-reasoning and DVK-memory, and the reasoned virtual knowledge affects generating responses.
Situation (1) is better than situation (4) can also reflect all of the modules play important roles in our model.
\textbf{In summary,} these modules of DMKCM designed for fusing structured and unstructured knowledge are helpful for response generating in terms of informativeness and coherence.
\begin{table}
\centering
\caption{Results of ablation study.}
\label{tab_abl}
\resizebox{\linewidth}{!}{ 
\renewcommand{\arraystretch}{0.8}
\begin{tabular}{clllllllll}
\toprule
\textbf{Dataset} & \textbf{Model} & \textbf{PPL} & \textbf{BLEU-1} & \textbf{BLEU-2} & \textbf{BLEU-3} & \textbf{BLEU-4} & \textbf{Dist-1} & \textbf{Dist-2} & \textbf{Flu./Inf./App.} \\
\midrule
\multirow{5}{*}{\textbf{PersonaChat}} 
& \textbf{DMKCM} & {\textbf{5.4549}} & {\textbf{0.5452}} & {\textbf{0.3149}} 
& {\textbf{0.1852}} & {\textbf{0.1087}} & {\textbf{0.0903}} & {\textbf{0.3328}} 
&{\textbf{1.62/1.43/1.35}} \\
 \cmidrule{2-10} 
 & - 2Hop &5.5568  & 0.4516 & 0.2608 &	0.1506  &  0.0870  & 0.0838 & 0.3233 
 &1.58/1.38/1.28 \\
 & - Mem and 2Hop & 6.1340  & 0.5005  & 0.2890 & 0.1669 & 0.0964 & 0.0782 & 0.2712 
 &1.53/1.31/1.18 \\
 & - 1Hop and Mem & 6.7423 & 0.4843 & 0.2796  & 0.1615 & 0.0933 &  0.0862 & 0.3134 
 & 1.52/1.40/1.26 \\
 & - 1Hop, Mem and 2Hop & 6.5172 & 0.5023 & 0.2900 & 0.1675 & 0.0967 & 0.0425 & 0.2239
 & 1.46/1.22/1.08 \\
\midrule\midrule
\multirow{5}{*}{\textbf{DailyDialog}} 
& \textbf{DMKCM} & {\textbf{5.7692}} & {\textbf{0.4928}} & {\textbf{0.2846}} & {\textbf{0.1645}} & {\textbf{0.0951}} & {\textbf{0.0801}} & {\textbf{0.3589}}
& {\textbf{1.58/1.48/1.44}} \\
\cmidrule{2-10}
  & - 2Hop & 7.6755 & 0.4643 & 0.2682 & 0.1550 & 0.0896 & 0.0744  & 0.3498 
  & 1.51/1.34/1.31 \\
 & - Mem and 2Hop & 8.8904 & 0.4632 & 0.2677 & 0.1547 & 0.0894 & 0.0683 & 0.3414 
 & 1.48/1.32/1.29 \\
 & - 1Hop and Mem & 8.0996 & 0.4572 & 0.2641 & 0.1526 & 0.0882 & 0.0792 & 0.3505 
 & 1.50/1.35/1.33 \\
  & - 1Hop, Mem and 2Hop & 10.1355 & 0.4205 & 0.2428 & 0.1403 & 0.0811 & 0.0633 & 0.3096 & 1.44/1.21/1.25 \\
\bottomrule
\end{tabular}
}
\end{table}
\subsection{Case Study}
\begin{figure}
\centering
\subfigure[A generated response example (knowledge from 1st hop)]{
\begin{minipage}[t]{0.45\linewidth}
\centering
\includegraphics[scale=0.23]{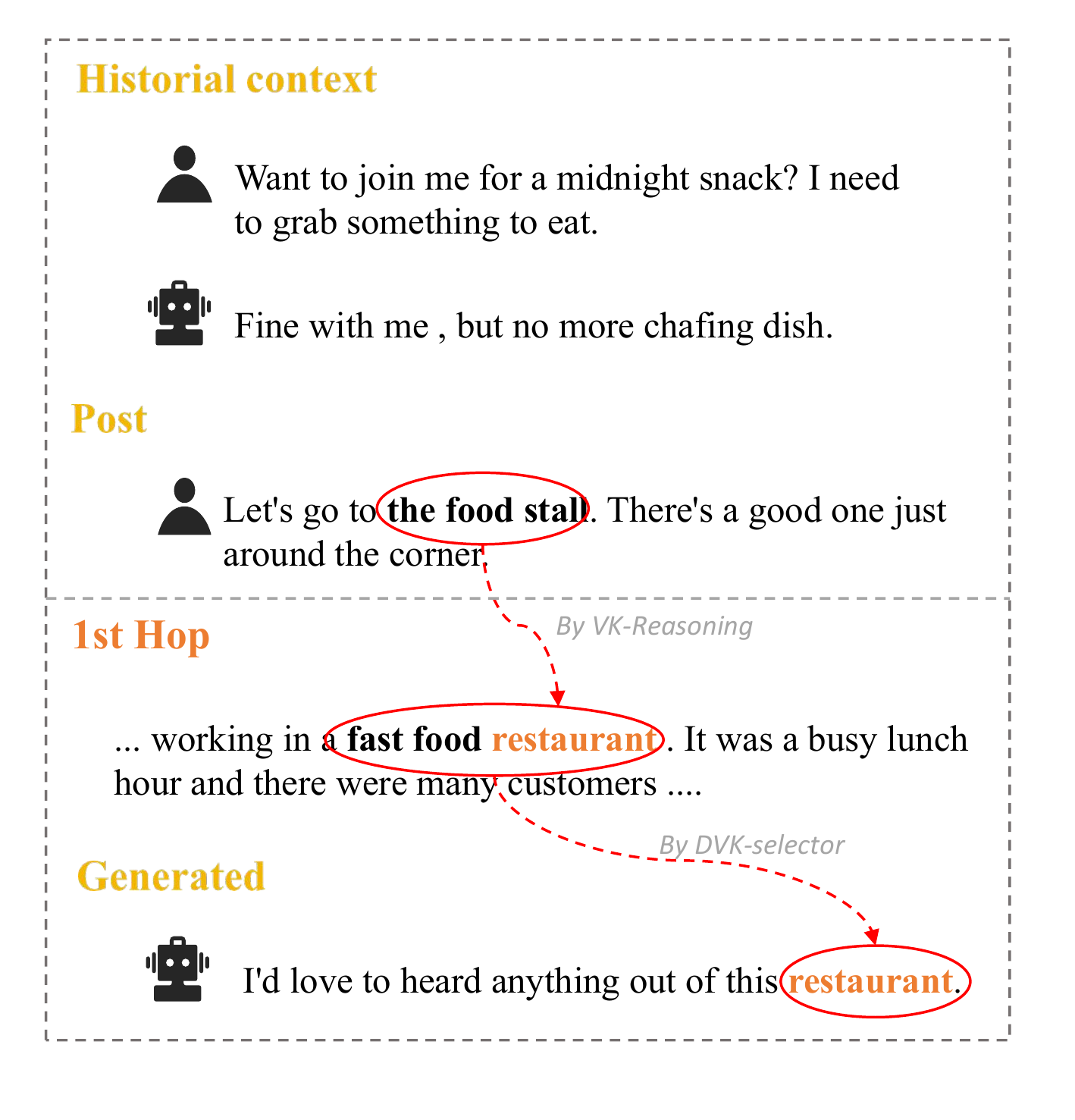} 
\end{minipage}
}
\subfigure[A generated response example (knowledge from history virtual knowledge and 2nd hop).]{
\begin{minipage}[t]{0.45\linewidth}
\centering
\includegraphics[scale=0.23]{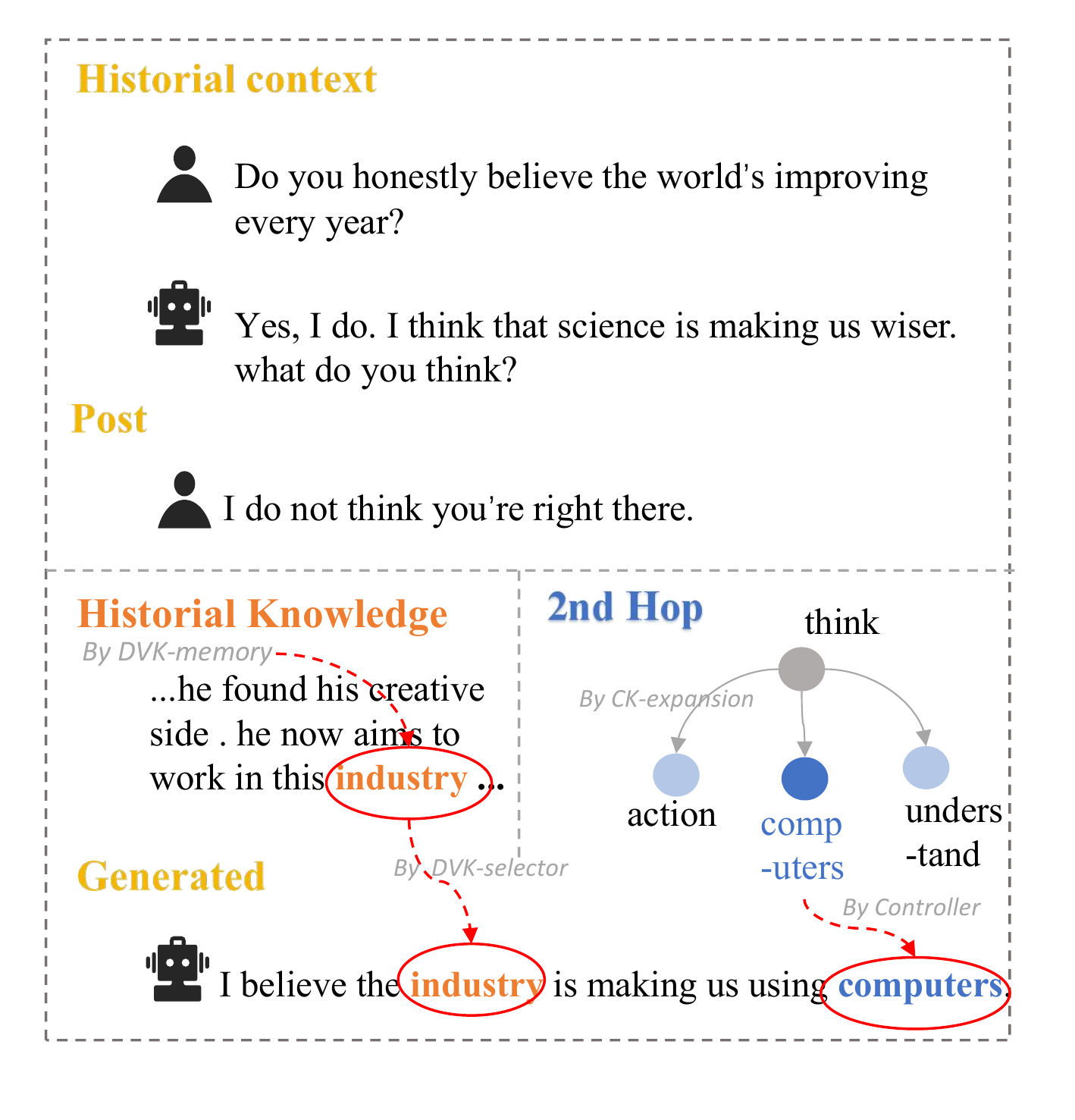}
\end{minipage}
}
\centering
\caption{Examples of DMKCM}
\label{fig_cases}
\end{figure}

Sample conversations are shown in Table~\ref{tab_cases} which indicate that DMKCM can generate better responses than the baselines on two conversation datasets. 
Traditional end-to-end models, e.g.,  Attn-S2S and Dir-VHRED, tend to generate simple, safe, and incoherent responses without access to knowledge. 
Despite knowledge-based model, e.g., CCM, GLKS, and MKST, which generate informative responses, still produces irrelevant responses.
Instead, DMKCM fuses knowledge from the knowledge graph and virtual knowledge as a whole and encodes more related information via DVK-selector and Controller modules, which supports generating more information and coherence response. 
In practice, the effect of these modules can be reflected, like Fig.~\ref{fig_cases}.
It thus generates more reasonable responses through better use of knowledge.
\begin{table}\large
\centering
\caption{Case Study of Generated Responses.} 
\label{tab_cases}
\resizebox{0.9\textwidth}{!}{ 
\renewcommand{\arraystretch}{1}
\begin{tabular}{|c|p{0.8\textwidth}|p{0.8\textwidth}|}
\hline
-
& \multicolumn{1}{c|}{\textbf{PersonaChat}} 
& \multicolumn{1}{c|}{\textbf{DailyDialog}} \\ \hline
-
& \begin{tabular}[c]{@{}p{0.8\textwidth}@{}}
\textbf{Historical Contexts}:\\ 
A1: Hey, how's it going? I am eating pizza.\\ 
B1: Not too bad. Pizza sounds great, I am on a diet.\\
\textbf{Post}:\\ 
A2: I'm a simple guy, I like pizza and trucks. But not diets.\end{tabular} & \begin{tabular}[c]{@{}p{0.8\textwidth}@{}}
\textbf{Historical Contexts}:\\ 
A1: Say, Jim, how about going for a few beers after dinner?\\ 
B1: You know that is tempting but is really not good for our fitness.\\
\textbf{Post}:\\ 
A2: What do you mean? It will help us to relax.\end{tabular} \\ \hline
\textbf{Attn-S2S}
& I love the walking. I like spicy food.
& Do you really think it? \\ \hline
\textbf{Transformer}
& That is cool. I like to eat healthy.
& \begin{tabular}[c]{@{}p{0.8\textwidth}@{}}
I am sorry, but i am not sure that it will be a good idea.\end{tabular} \\ \hline
\textbf{Dir-VHRED}
& \begin{tabular}[c]{@{}p{0.8\textwidth}@{}}
I am always looking for ways to make money, I can make some money.\end{tabular}
& \begin{tabular}[c]{@{}p{0.8\textwidth}@{}}
But you know, it's not so much. Now that you have to do the drying up. \end{tabular}\\ \hline
\textbf{GLKS} & 
\begin{tabular}[c]{@{}p{0.8\textwidth}@{}}
I love food too much. I am trying to open a restaurant next day.\end{tabular}
& Do you really think so? \\ \hline
\textbf{CCM} & 
I am a good aggressive. I am a good teacher.
& I am like to get harvesting of the day. \\ \hline
\textbf{MKST} & 
I like to work out.
& \multicolumn{1}{c|}{-} \\ \hline
\textbf{DMKCM} & 
\begin{tabular}[c]{@{}p{0.8\textwidth}@{}}
I am a waitress at a small restaurant. I am a vegan, but I like pizza.\end{tabular}
& \begin{tabular}[c]{@{}p{0.8\textwidth}@{}}
Do you think so? I do not really think we should get looking forward to it, it will just make us fat.\end{tabular} \\ \hline
\end{tabular}
}
\end{table}
\section{Conclusion}
To solve the challenge of lacking informative response in multi-turn dialogue generation, we propose a novel model, DMKCM.
The existing methods of introducing knowledge into dialogue generation have some limits, so we combine virtual KB and commonsense KG to help generate better responses.
In addition,  we find that history virtual knowledge can improve responses and provide a new dynamically delayed updating strategy to store and filter history virtual knowledge.
Experimental results on two datasets show that DMKCM can generate a more informative dialog with appropriate content ordering.

%
%
%
%
%
\bibliographystyle{splncs04}
%

\end{document}